\documentclass[a4paper,conference]{IEEEtran}
\IEEEoverridecommandlockouts
\usepackage[numbers]{natbib}
\usepackage{amsmath,amssymb,amsfonts}
\usepackage{algorithmic}
\usepackage{graphicx}
\usepackage{textcomp}
\usepackage{xcolor}
\usepackage[breaklinks=true]{hyperref}  
\usepackage{cleveref}
\usepackage[utf8]{inputenc}
\usepackage[T1]{fontenc}
\usepackage{soul}
\usepackage{multirow}
\usepackage{tikz}
\usepackage{url}

\usepackage[numbers]{natbib} 
\renewcommand{\refname}{\MakeUppercase{References}} 

\begin{document}

\title{\fontsize{19pt}{22pt}\selectfont\bfseries
\uppercase{Combinations of Fast Activation and Trigonometric Functions in Kolmogorov-Arnold Networks}\\
}

\author{\IEEEauthorblockN{Hoang-Thang Ta}
\IEEEauthorblockA{\textit{Faculty of Information Technology} \\
\textit{Dalat University}\\
Dalat, Vietnam \\
thangth@dlu.edu.vn}
\and
\IEEEauthorblockN{Duy-Quy Thai}
\IEEEauthorblockA{\textit{Faculty of Information Technology} \\
\textit{Dalat University}\\
Dalat, Vietnam \\
quytd@dlu.edu.vn}
\and
\IEEEauthorblockN{ Phuong-Linh Tran-Thi}
\IEEEauthorblockA{\textit{Faculty of Information Technology} \\
\textit{Dalat University}\\
Dalat, Vietnam \\
linhttp@dlu.edu.vn}
}

\maketitle

\begin{abstract}
For years, many neural networks have been developed based on the Kolmogorov-Arnold Representation Theorem (KART), which was created to address Hilbert's 13th problem. Recently, relying on KART, Kolmogorov-Arnold Networks (KANs) have attracted attention from the research community, stimulating the use of polynomial functions such as B-splines and RBFs. However, these functions are not fully supported by GPU devices and are still considered less popular. In this paper, we propose the use of fast computational functions, such as ReLU and trigonometric functions (e.g., ReLU, sin, cos, arctan), as basis components in Kolmogorov-Arnold Networks (KANs). By integrating these function combinations into the network structure, we aim to enhance computational efficiency. Experimental results show that these combinations maintain competitive performance while offering potential improvements in training time and generalization. We provide our code repo at:  \url{https://github.com/hoangthangta/FC_KAN}.
\end{abstract}

\begin{IEEEkeywords}
Kolmogorov-Arnold Networks, function combinations, activation functions, trigonometric functions
\end{IEEEkeywords}

\section{\uppercase{Introduction}}

\citet{liu2024kan,liu2024kan2.0} proposed a novel neural architecture called Kolmogorov-Arnold Networks (KANs), inspired by the Kolmogorov-Arnold representation theorem (KART). Their work highlights the use of learnable functions, specifically B-splines, as "edges" to better fit training data, contrasting with traditional multi-layer perceptrons (MLPs), which use fixed activation functions at the "nodes". Subsequent studies have explored replacing B-splines with other basis and polynomial functions~\cite{li2024kolmogorov,athanasios2024,bozorgasl2024wav,ss2024chebyshev}. 

In another line of research, several studies have investigated function combinations within KANs~\cite{ta2024bsrbf,yang2024activation,altarabichi2024rethinking,liu2024kan2.0}. In MLP-based architectures, combining multiple activation functions has also been explored to improve model performance and stability~\cite{jie2021regularized,xu2020comparison,zhang2015genetic}. In the context of KANs, these combinations are applied to the outer functions, rather than to activation functions. For instance, \citet{ta2024bsrbf} introduced BSRBF-KAN, which integrates B-splines and Gaussian Radial Basis Functions (GRBFs) across all layers. \citet{liu2024kan2.0} proposed MultKAN, employing element-wise multiplication and addition to better capture data patterns in small-scale examples. Moreover, \citet{ta2024fc} introduced FC-KAN, where function combinations are applied at the output level under the assumption of low-dimensional data representations. However, their approach still faces challenges related to long training times and parameter scalability.

In this paper, we take advantage of the fast computation speed of activation functions like ReLU and trigonometric functions such as sin, cos, and arctan. We combine these functions using simple methods such as sum and product on low-dimensional data in FC-KAN~\cite{ta2024fc}. More complex combinations, such as the addition of sum and product, or quadratic and cubic function representations are not chosen because they increase training time. Experiments on image classification tasks using MNIST and Fashion-MNIST show that our method effectively reduces training time compared to the use of B-splines and RBFs, while achieving competitive performance with MLPs and other KANs.

\section{\uppercase{Related Works}}\label{sec_related_works}

Kolmogorov’s 1957 solution to Hilbert’s 13th problem introduced the Kolmogorov-Arnold representation theorem (KART)~\cite{kolmogorov1957representation,braun2009constructive}, proving that any multivariate continuous function can be expressed using single-variable functions and additions. This concept gained renewed interest with the proposal of Kolmogorov-Arnold Networks (KANs)~\cite{liu2024kan}, leading to applications in expensive optimization~\cite{hao2024first}, keyword spotting~\cite{xu2024effective}, mechanics~\cite{abueidda2024deepokan}, quantum computing~\cite{kundu2024kanqas,wakaura2024variational,troy2024sparks}, survival analysis~\cite{knottenbelt2024coxkan}, time series forecasting~\cite{genet2024tkan,xu2024kolmogorov,vaca2024kolmogorov,genet2024temporal,han2024kan4tsf,xu2024kan4drift}, and computer vision~\cite{li2024u,cheon2024demonstrating,ge2024tc}.

KAN variants have utilized well-known mathematical functions to enhance model expressiveness and performance. These include B-splines~\cite{de1972calculating,liu2024kan}, Gaussian radial basis functions (GRBFs)~\cite{li2024kolmogorov,abueidda2024deepokan,ta2024bsrbf}, reflection switch activation functions (RSWAF)~\cite{athanasios2024}, Chebyshev and Legendre polynomials~\cite{torchkan,ss2024chebyshev}, Fourier transforms~\cite{xu2024fourierkan}, wavelets~\cite{bozorgasl2024wav,seydi2024unveiling}, and other polynomial functions~\cite{teymoor2024exploring}.

Several recent studies have explored function combinations to design improved KAN architectures. BSRBF-KAN~\cite{ta2024bsrbf} combined B-splines and RBFs to improve convergence on MNIST and Fashion-MNIST. MultKAN~\cite{liu2024kan2.0} introduced multiplication nodes to better capture multiplicative structures but focused only on small-scale examples. S-KAN~\cite{yang2024activation} used adaptive strategies to combine activation functions at each node. DropKAN~\cite{altarabichi2024rethinking,altarabichi2024dropkan} replaced summation with averaging in neurons to boost performance and stability. ReLU-KAN~\cite{qiu2024relu} used ReLU-based basis functions with matrix operations for better parallelization. Lastly, FC-KAN~\cite{ta2024fc} combined B-splines and RBFs using various methods on low-dimensional data.

\section{\uppercase{Methodology}}
\subsection{KART and KAN}
\begin{figure*}[tp]
  \centering
\includegraphics[scale=0.95]{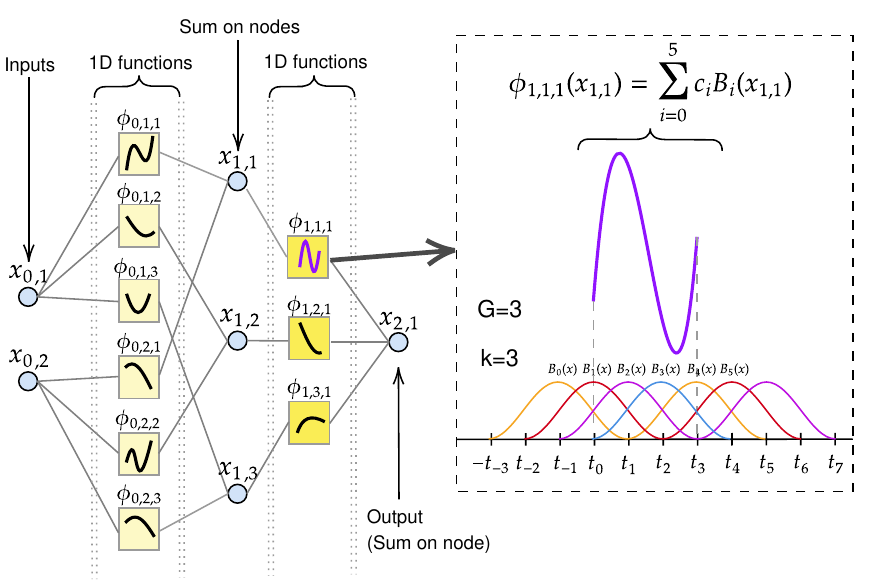}
  \centering
  \caption{[Left]: the architecture of KAN(2,3,1). [Right]: a showcase of computing \(\phi_{1,1,1}\) using control points and B-splines, where \(G\) and \(k\) denote the grid size and spline order, respectively, with the total number of B-splines equal to \(G + k = 3 + 3 = 6\)~\cite{ta2024fc}.}
\label{fig:kan_diagram}
\end{figure*}

Kolmogorov-Arnold Representation Theorem proves that any continuous multivariate function \( f \), defined over a bounded domain, can be represented using a finite combination of continuous univariate functions along with addition operations~\cite{chernov2020gaussian,schmidt2021kolmogorov}. Given a set of variables \( \mathbf{x} = \{x_1, x_2, \ldots, x_n\} \), where \( n \) denotes the number of variables, the continuous multivariate function \( f(\mathbf{x}) \) can be formulated as:

\begin{equation}
\begin{aligned}
f(\mathbf{x}) = f(x_1, \ldots, x_n) = \sum_{q=1}^{2n+1} \Phi_q \left( \sum_{p=1}^{n} \phi_{q,p}(x_p) \right) 
\end{aligned}
\label{eq:kart}
\end{equation}
This representation involves two types of summations: an outer sum and an inner sum. The outer sum, \( \sum_{q=1}^{2n+1} \), combines \( 2n+1 \) terms of the form \( \Phi_q \colon \mathbb{R} \to \mathbb{R} \). The inner sum, in contrast, aggregates \( n \) terms for each index \( q \), where each term \( \phi_{q,p} \colon [0,1] \to \mathbb{R} \) is a continuous function of a single variable \( x_p \).

First, a MLP is a classical neural network architecture that utilizes affine transformations followed by non-linear activation functions. An input \( \mathbf{x} \) is passed via a sequence of layers (from layer \( 0 \) to layer \( L-1 \)) by applying weight matrices and biases, followed by a non-linear activation function \( \sigma \), to generate the final output as:

\begin{equation}
\begin{aligned}
\text{MLP}(\mathbf{x}) &= (W_{L-1} \circ \sigma \circ W_{L-2} \circ \sigma \circ \cdots \circ W_1 \circ \sigma \circ W_0) \mathbf{x} \\
&= \sigma \left( W_{L-1} \sigma \left( W_{L-2} \sigma \left( \cdots \sigma \left( W_1 \sigma \left( W_0 \mathbf{x} \right) \right) \right) \right) \right)
\end{aligned}
\label{eq:mlp}
\end{equation}

When building KANs, \citet{liu2024kan} suggested increasing both the width and depth of the network by searching for appropriate functions \( \Phi_q \) and \( \phi_{q,p} \) in \Cref{eq:kart}. A KAN with \( L \) layers processes the input \( \mathbf{x} \) to generate the output as:

\begin{equation}
\begin{aligned}
\text{KAN}(\mathbf{x}) = (\Phi_{L-1} \circ \Phi_{L-2} \circ \cdots \circ \Phi_1 \circ \Phi_0)\mathbf{x}
\end{aligned}
\label{eq:kan}
\end{equation}
Here, \( \Phi_{l} \) denotes the function matrix of the \( l^{\text{th}} \) KAN layer, representing a set of pre-activation functions. Let us denote the \( i^{\text{th}} \) neuron in layer \( l \) and the \( j^{\text{th}} \) neuron in layer \( l+1 \). The activation function \( \phi_{l,i,j} \) connects neuron \( (l, i) \) to neuron \( (l+1, j) \):

\begin{equation}
\begin{aligned}
\phi_{l,j,i}, \quad l = 0, \cdots, L - 1, \quad i = 1, \cdots, n_l, \quad j = 1, \cdots, n_{l+1}
\end{aligned}
\label{eq:acti_funct}
\end{equation}
Let \( n_l \) denote the number of neurons in the \( l^{\text{th}} \) layer. Accordingly, the function matrix \( \Phi_{l} \) can be represented as an \( n_{l+1} \times n_l \) matrix of activation functions as follows:

\begin{equation}
\begin{aligned}
\mathbf{x}_{l+1} = 
\underbrace{\left(
\begin{array}{cccc}
\phi_{l,1,1}(\cdot) & \phi_{l,1,2}(\cdot) & \cdots & \phi_{l,1,n_l}(\cdot) \\
\phi_{l,2,1}(\cdot) & \phi_{l,2,2}(\cdot) & \cdots & \phi_{l,2,n_l}(\cdot) \\
\vdots & \vdots & \ddots & \vdots \\
\phi_{l,n_{l+1},1}(\cdot) & \phi_{l,n_{l+1},2}(\cdot) & \cdots & \phi_{l,n_{l+1},n_l}(\cdot)
\end{array}\right)}_{\Phi_{l}} \mathbf{x}_l
\label{eq:function_matrix}
\end{aligned}
\end{equation}

\Cref{fig:kan_diagram} shows the architecture of KAN(2,3,1) with 2 input nodes, passing through 3 hidden nodes, and producing 1 output. The figure also provides the formula for calculating one-dimensional functions such as B-splines.

\subsection{Function combination}
\citet{ta2024fc} introduced FC-KAN (Function Combinations in Kolmogorov-Arnold Networks), which combines popular functions originally used in KANs, such as B-splines, RBFs, and DoGs (Derivatives of Gaussians). While achieving strong performance during training, FC-KAN still presents certain limitations, particularly the overuse of parameters and extended training times. To address this, we propose using fast functions such as ReLU, sin, cos, and arctan as replacements. We chose ReLU because it is simple and computationally efficient compared to other activation functions. We selected sin, cos, and arctan because they are faster to compute than some other trigonometric functions.

Given an input \(\mathbf{x}\), its output \(o\) is passed through the ReLU, sine, cosine, and arctangent functions as follows:

\begin{equation}
\begin{aligned}
\label{eq:output_combi}
\mathbf{o}_{ReLU} &= \max(0, \mathbf{x}), \\
\mathbf{o}_{sin} &= \sin(\mathbf{x}), \\
\mathbf{o}_{cos} &= \cos(\mathbf{x}), \\
\mathbf{o}_{arctan} &= \arctan(\mathbf{x}).
\end{aligned}
\end{equation}

Next, we rely on element-wise operations, including sum (\Cref{eq:output_sum}) and product (\Cref{eq:output_product}), to generate the output. Given an input \(\mathbf{x}\) and a set of functions \( F = \{f_1, f_2, \dots, f_n\} \), where \( n \) is the number of functions used, the input \(\mathbf{x}\) is independently passed through each function \( f_i \) via the network layers, producing the output \(\mathbf{o}_i\) as:

\begin{subequations}
\begin{equation}
  \mathbf{o}_{sum} = \sum_{i=1}^{n}\mathbf{o}_i  = \mathbf{o}_1 + \mathbf{o}_2 + \cdots + \mathbf{o}_n
  \label{eq:output_sum}
\end{equation}    
\begin{equation}
  \mathbf{o}_{prod} = \bigodot_{i=1}^{n}\mathbf{o}_i = \mathbf{o}_1 \odot \mathbf{o}_2 \odot \cdots \odot \mathbf{o}_n
  \label{eq:output_product}
\end{equation}
\end{subequations}

\section{Experiments}

\subsection{Training configuration}
\label{train_config}

\begin{table*}[t]
	\caption{\uppercase{Model structures and parameters used.}}
	\centering
	\begin{tabular}{p{4cm}p{3.5cm}p{3cm}p{2cm}}
            \hline
		\textbf{Dataset} & \textbf{Model} & \textbf{Network structure} & \textbf{\#Params} \\
    \hline 
    \multirow{6}{4cm}{\textbf{MNIST + Fashion-MNIST}}  &  BSRBF-KAN	& (784, 64, 10) & 459024  \\
&  FastKAN	& (784, 64, 10) & 459098 \\
& FasterKAN	& (784, 64, 10) & 408206 \\
& EfficientKAN	& (784, 64, 10) & \textbf{508160} \\
& FC-KAN* &	(784, 64, 10) & \textbf{52496} \\
& MLP & (784, 64, 10) & 52512 \\
            \hline
            \multicolumn{4}{l}{*All variants of FC-KAN: sin+cos, sin+ReLU, sin+arctan, cos+ReLU, cos+arctan, arctan+ReLU}  \\
             \hline
	\end{tabular}
	\label{tab:model_params}
\end{table*}

For each model, we conducted 3 independent training runs on the MNIST~\cite{deng2012mnist} and Fashion-MNIST~\cite{xiao2017fashion} datasets to ensure a more reliable evaluation of overall performance. We computed the average across all runs to reduce the effects of training variability and to more accurately estimate the models' maximum potential. To keep the network design straightforward, all models-BSRBF-KAN~\cite{ta2024bsrbf}, EfficientKAN~\cite{Blealtan2024}, FastKAN~\cite{li2024kolmogorov}, FasterKAN~\cite{athanasios2024}, FC-KAN, and MLP-used only activation functions (ReLU), linear transformations, and layer normalization. In our experiments, FC-KAN still lags significantly behind several popular neural networks such as CNNs; therefore, we decided not to include those results in the paper.

As presented in \Cref{tab:model_params}, each model employs a network architecture of \((784, 64, 10)\), consisting of 784 input neurons, 64 hidden neurons, and 10 output neurons corresponding to the 10 classes (digits 0-9). The number of parameters in FC-KAN is comparable to that of MLPs and significantly fewer than in the other models. Training was conducted for 25 epochs on MNIST and 35 epochs on Fashion-MNIST. With these epoch counts, most KAN variants begin to converge, achieving training accuracies of nearly 100\%.

Other hyperparameters were kept consistent across all models: \texttt{batch\_size=64}, the \texttt{AdamW} optimizer, \texttt{CrossEntropy} loss function, an \texttt{ExponentialLR} scheduler with a learning rate of \(1 \times 10^{-3}\) and \(\gamma = 0.8\), and a weight decay of \(1 \times 10^{-4}\). These choices reflect standard practice for training on MNIST and Fashion-MNIST, making extensive hyperparameter tuning unnecessary.  In the FC-KAN models, we combined 2 out of 4 functions-sin, cos, arctan, and ReLU-to generate 6 variants: FC-KAN (sin+cos), FC-KAN (sin+arctan), FC-KAN (sin+ReLU), FC-KAN (cos+ReLU), FC-KAN (arctan+cos), and FC-KAN (arctan+ReLU).

\subsection{Function speed}
\begin{figure*}[tp]
  \centering
 \includegraphics[width=0.85\textwidth]{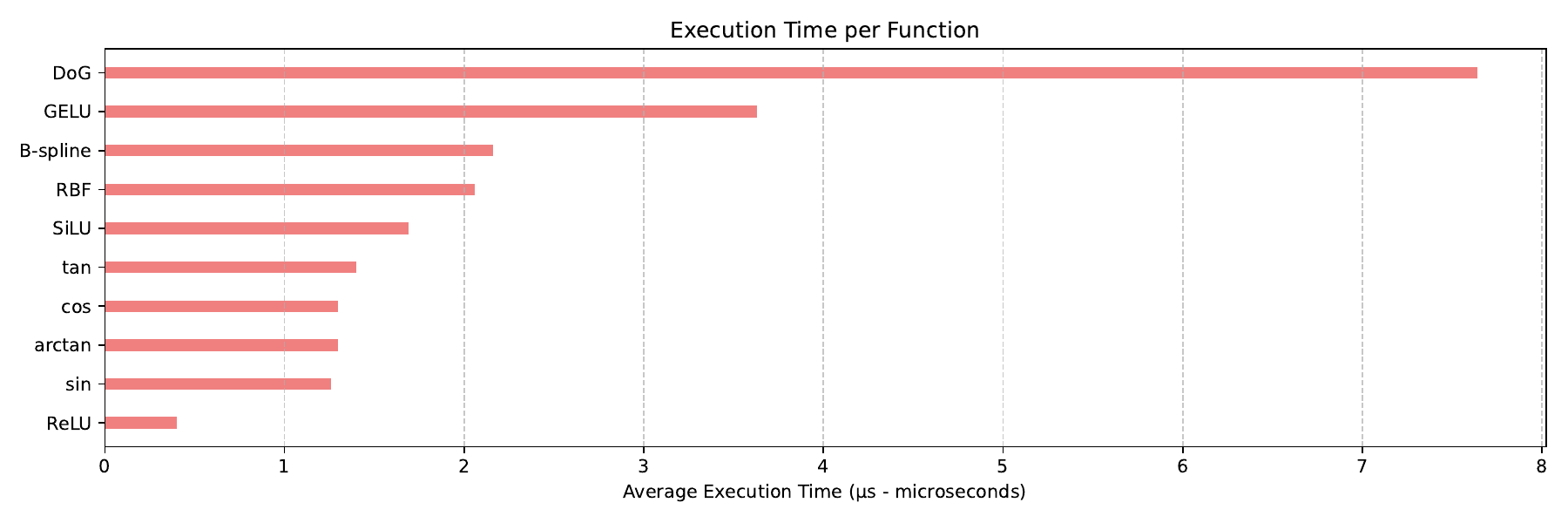}
  \centering
  \caption{Average execution time (\(\mu\)s -- microseconds) for processing 1,000,000 inputs per function.}
\label{fig:func_speed}
\end{figure*}

In this section, we conduct an experiment to evaluate the computational speed of various functions-including B-spline, RBF, DoG, ReLU, sin, cos, tan, and arctan, when processing a set of inputs. We create an input array of size 1,000,000, pass each element through these functions, and then calculate the average execution time, converted to microseconds (\(\mu s\)). As shown in \Cref{fig:func_speed}, the B-spline function takes the longest time to execute, while ReLU is approximately twice as fast, followed by sin, cos, and arctan. This experiment illustrates the rationale behind selecting these functions for the combinations used in FC-KAN.

\subsection{Model performance}
In this section, we evaluate various function combinations in FC-KAN and compare their performance, as presented in \Cref{tab:fc_kan_combi}. On the MNIST dataset, the use of \texttt{cos} alone results in the shortest training time. The \texttt{sin+cos (product)} combination achieves the highest training accuracy, while \texttt{sin+cos (sum)} delivers the best validation accuracy and F1 score. For Fashion-MNIST, the \texttt{sin} function alone leads in both training speed and training accuracy. In terms of generalization, the \texttt{sin+arctan (product)} combination achieves the highest validation accuracy and F1 score, demonstrating its effectiveness on more complex data. 

\begin{table*}[t]
    \caption{The average metric values in 3 training runs on MNIST and Fashion-MNIST of FC-KAN models.}
    \centering
    \begin{tabular}{p{2cm}p{3.5cm}p{2.5cm}p{2.5cm}p{2.5cm}p{2cm}}
        \hline
        \textbf{Dataset} & \textbf{Model} & \textbf{Train. Acc.} & \textbf{Val. Acc.} & \textbf{F1} & \textbf{Time (s)} \\
        \hline
        \multirow{20}{2cm}{\textbf{MNIST}} 
        & arctan+ReLU (product) & 99.83 ± 0.04 & 97.43 ± 0.11 & 97.40 ± 0.11 & 191.12 \\
        & arctan+ReLU (sum) & 99.21 ± 0.10 & 97.58 ± 0.01 & 97.55 ± 0.01 & 189.82 \\
        & arctan & 98.31 ± 0.02 & 96.97 ± 0.07 & 96.94 ± 0.08 & 166.26 \\
        & ReLU & 99.53 ± 0.01 & 97.60 ± 0.02 & 97.58 ± 0.02 & 187.25 \\
        & cos+arctan (product) & 99.10 ± 0.06 & 97.50 ± 0.01 & 97.47 ± 0.01 & 195.96 \\
        & cos+arctan (sum) & 99.32 ± 0.01 & 97.57 ± 0.02 & 97.53 ± 0.02 & 190.84 \\
        & cos+ReLU (product) & 99.88 ± 0.05 & 97.60 ± 0.06 & 97.57 ± 0.06 & 196.04 \\
        & cos+ReLU (sum) & 98.98 ± 0.03 & 97.22 ± 0.03 & 97.18 ± 0.03 & 187.72 \\
        & cos & 99.48 ± 0.01 & 97.59 ± 0.08 & 97.55 ± 0.08 & \textbf{165.88} \\
        & sin+arctan (product) & 99.81 ± 0.09 & 97.51 ± 0.03 & 97.48 ± 0.03 & 193.49 \\
        & sin+arctan (sum) & 99.59 ± 0.13 & 97.56 ± 0.03 & 97.52 ± 0.03 & 189.77 \\
        & sin+ReLU (product) & 99.80 ± 0.11 & 97.60 ± 0.07 & 97.57 ± 0.08 & 191.83 \\
        & sin+ReLU (sum) & 99.70 ± 0.07 & 97.58 ± 0.03 & 97.55 ± 0.03 & 183.53 \\
        & sin+cos (product) & \textbf{99.90 ± 0.01} & 97.28 ± 0.05 & 97.25 ± 0.06 & 194.90 \\
        & sin+cos (sum) & 99.66 ± 0.02 & \textbf{97.64 ± 0.07} & \textbf{97.62 ± 0.07} & 188.80 \\
        & sin & 99.82 ± 0.05 & 97.32 ± 0.03 & 97.29 ± 0.03 & 166.08 \\
        \hline
        \multirow{20}{2cm}{\textbf{Fashion-MNIST}} 
        & arctan+ReLU (product) & 92.52 ± 0.06 & 88.48 ± 0.02 & 88.42 ± 0.02 & 266.40 \\
        & arctan+ReLU (sum) & 92.49 ± 0.05 & 88.36 ± 0.10 & 88.31 ± 0.10 & 256.23 \\
        & arctan & 91.78 ± 0.10 & 88.52 ± 0.07 & 88.42 ± 0.06 & 236.62 \\
        & ReLU & 93.13 ± 0.01 & 88.14 ± 0.09 & 88.08 ± 0.09 & 257.17 \\
        & cos+arctan (product) & 92.75 ± 0.05 & 88.70 ± 0.05 & 88.63 ± 0.06 & 264.84 \\
        & cos+arctan (sum) & 93.62 ± 0.08 & 88.66 ± 0.05 & 88.55 ± 0.06 & 264.33 \\
        & cos+ReLU (product) & 93.54 ± 0.06 & 88.05 ± 0.11 & 88.03 ± 0.12 & 267.97 \\
        & cos+ReLU (sum) & 92.96 ± 0.05 & 88.11 ± 0.07 & 88.05 ± 0.08 & 257.83 \\
        & cos & 95.17 ± 0.04 & 88.09 ± 0.09 & 88.03 ± 0.07 & 237.24 \\
        & sin+arctan (product) & 94.06 ± 0.08 & \textbf{89.38 ± 0.02} & \textbf{89.30 ± 0.02} & 267.24 \\
        & sin+arctan (sum) & 93.94 ± 0.06 & 89.07 ± 0.09 & 88.98 ± 0.10 & 260.48 \\
        & sin+ReLU (product) & 94.36 ± 0.07 & 89.30 ± 0.18 & 89.23 ± 0.17 & 268.30 \\
        & sin+ReLU (sum) & 93.70 ± 0.19 & 88.97 ± 0.14 & 88.88 ± 0.13 & 262.13 \\
        & sin+cos (product) & 94.32 ± 0.24 & 89.12 ± 0.15 & 89.04 ± 0.14 & 259.68 \\
        & sin+cos (sum) & 94.71 ± 0.11 & 89.08 ± 0.12 & 89.02 ± 0.12 & 258.28 \\
        & sin & \textbf{94.76 ± 0.01} & 89.13 ± 0.07 & 89.08 ± 0.09 & \textbf{233.51} \\
        \hline
        \multicolumn{6}{l}{Train. Acc = Training Accuracy, Val. Acc. = Validation Accuracy} \\
        \hline
    \end{tabular}
    \label{tab:fc_kan_combi}
\end{table*}


We then select the two best-performing FC-KAN models—\texttt{sin+cos (sum)} and \texttt{sin+arctan (product)}—for comparison against MLP and other KAN variants, as presented in \Cref{tab:fc_kan_vs_others}. On the MNIST dataset, the \texttt{sin+cos (sum)} model slightly outperforms other KAN-based models and is marginally behind the MLP in terms of validation accuracy, but it incurs a 15.26\% increase in training time compared to MLP. On the Fashion-MNIST dataset, the \texttt{sin+arctan (product)} model performs better than the MLP but lags behind the other KAN variants. Its training time is shorter than that of EfficientKAN and BSRBF-KAN but remains higher than the rest. These results suggest that further optimization is needed to reduce the training time of FC-KAN models while maintaining competitive performance.

\begin{table*}[t]
    \caption{The average metric values in 3 training runs on MNIST and Fashion-MNIST of FC-KAN models.}
    \centering
    \begin{tabular}{p{2cm}p{3.5cm}p{2.5cm}p{2.5cm}p{2.5cm}p{2cm}}
        \hline
        \textbf{Dataset} & \textbf{Model} & \textbf{Train. Acc.} & \textbf{Val. Acc.} & \textbf{F1} & \textbf{Time (s)} \\
        \hline
\multirow{6}{2cm}{\textbf{MNIST}} 
& sin+cos (sum) & 99.66 ± 0.02 & 97.64 ± 0.07 & 97.62 ± 0.07 & 188.80 \\ 
& BSRBF-KAN & \textbf{100.00 ± 0.00} & 97.63 ± 0.02 & 97.59 ± 0.02 & 222.57 \\
& FastKAN & 99.98 ± 0.00 & 97.56 ± 0.09 & 97.52 ± 0.10 & 184.81 \\
& FasterKAN & 98.69 ± 0.03 & 97.65 ± 0.06 & 97.62 ± 0.07 & 177.14 \\
& MLP & 99.78 ± 0.13 & \textbf{97.69 ± 0.02} & \textbf{97.66 ± 0.02} & \textbf{163.80} \\
& EfficientKAN & 99.45 ± 0.03 & 97.39 ± 0.01 & 97.35 ± 0.01 & 191.51 \\
\hline
\multirow{6}{2cm}{\textbf{Fashion-MNIST}} 
& sin-arctan (product) & 94.06 ± 0.08 & 89.38 ± 0.02 & 89.30 ± 0.02 & 267.24 \\ 
& BSRBF-KAN & \textbf{99.33 ± 0.07} & 89.51 ± 0.14 & 89.49 ± 0.15 & 310.19 \\ 
& FastKAN & 98.26 ± 0.04 & \textbf{89.57 ± 0.08} & \textbf{89.54 ± 0.08} & 262.43 \\ 
& FasterKAN & 94.58 ± 0.05 & 89.39 ± 0.05 & 89.35 ± 0.03 & 252.38 \\ 
& MLP & 93.73 ± 0.36 & 89.10 ± 0.14 & 89.02 ± 0.14 & \textbf{231.25} \\ 
& EfficientKAN & 94.99 ± 0.02 & 89.07 ± 0.08 & 89.01 ± 0.09 & 277.02 \\
\hline
        \multicolumn{6}{l}{Train. Acc = Training Accuracy, Val. Acc. = Validation Accuracy} \\
        \hline
\end{tabular}
\label{tab:fc_kan_vs_others}
\end{table*}

\section{\uppercase{Limitation}}
Our study is limited to experiments on small, monochannel datasets such as MNIST and Fashion-MNIST. Moreover, we only apply a shallow network architecture (784, 64, 10), so the performance of deeper structures remains unknown. Another limitation is the lack of a solid mathematical foundation explaining why these function combinations better capture data features compared to single functions. Moreover, there are other fast functions we have yet to explore, and we have not compared FC-KAN to other modern neural networks. In summary, further experiments and more ablation studies are needed to clarify the strengths of FC-KAN and identify opportunities for improvement.

\section{\uppercase{Conclusion}}
We introduced combinations of activation and trigonometric functions such as ReLU, sine, cosine, and arctan to capture data features in Kolmogorov–Arnold Networks. These combination methods operate on low-dimensional data within network layers, merging function outputs using two approaches: sum and product. Experiments show that our FC-KAN, which uses activation and trigonometric functions, achieves competitive performance compared to MLPs and other KAN variants while maintaining the same number of parameters as MLPs. 

Our work still has some problems, notably that the combinations of fast functions, such as activation and trigonometric functions, in FC-KAN do not show significant superiority compared to other KANs. Furthermore, the use of these functions diverges from the traditional boundary of KANs, which typically rely on B-splines defined by parameters such as grid size and spline order.

In future work, we aim to address these remaining challenges and explore additional functions and combination methods to further optimize FC-KAN and validate its effectiveness on real-world datasets. We also consider the approach by \citet{qiu2024relu}, who use ReLU to mimic the behavior of B-splines.

\section*{\uppercase{Acknowledgment}}
This research is funded by the Foundation for Science and Technology Development of Dalat University, according to Decision No. 1153/QĐ-ĐHDL, dated on September 30, 2024.

{\fontsize{8}{9.6}\selectfont
\bibliographystyle{IEEEtranN}

\bibliography{bibi}  
}

\end{document}